%% file: iclr2021_conference.tex
\documentclass{article} 
\usepackage{iclr2021_conference,times}

\input{math_commands.tex}

\usepackage{hyperref}
\usepackage{url}

\usepackage{amsthm}
\newtheorem{myproposition}{Proposition}
\usepackage{graphicx}
\usepackage{amsmath}
\usepackage{color}
\usepackage{url}
\usepackage{algorithm}
\usepackage{algorithmic}
\usepackage{natbib}

\usepackage{subcaption}
\usepackage[switch,displaymath]{lineno}

\makeatletter
\let\LN@align\align
\let\LN@endalign\endalign
\renewcommand{\align}{\linenomath\LN@align}
\renewcommand{\endalign}{\LN@endalign\endlinenomath}
\let\LN@gather\gather
\let\LN@endgather\endgather
\renewcommand{\gather}{\linenomath\LN@gather}
\renewcommand{\endgather}{\LN@endgather\endlinenomath}
\makeatother

\title{Towards Robustness Against \\ Natural Language 
Word Substitutions}

\author{Xinshuai Dong  \\
Nanyang Technological University, Singapore \\
\texttt{dongxinshuai@outlook.com} \\
\And
Anh Tuan Luu \\
Nanyang Technological University, Singapore \\
\texttt{anhtuan.luu@ntu.edu.sg} \\
\AND
Rongrong Ji 
\\
Xiamen University, China  {~~~~~~~~~~~~~~~~~~~~~~} \\
\texttt{rrji@xmu.edu.cn}\\
\And
Hong Liu \\
National Institute of Informatics, Japan \\
\texttt{hliu@nii.ac.jp}
}

\iclrfinalcopy 
\begin{document}

\maketitle

\begin{abstract}

   Robustness against word substitutions has a well-defined and widely acceptable form, \emph{i.e.}, using semantically similar words as substitutions, and thus it is considered as a fundamental stepping-stone towards broader robustness in natural language processing. 
   Previous defense methods capture word substitutions in vector space by using either $l_2$-ball or hyper-rectangle, which results in perturbation sets that are not inclusive enough or unnecessarily large, 
   and thus impedes mimicry of worst cases for robust training.
   In this paper, we introduce a novel \textit{Adversarial Sparse Convex Combination} (ASCC) method.
   We model the word substitution attack space as a convex hull and leverages a regularization term to enforce perturbation towards an actual substitution,
    thus aligning our modeling better with the discrete textual space.
   Based on the ASCC method, we further propose ASCC-defense,
    which leverages ASCC to generate worst-case perturbations and incorporates adversarial training towards robustness.
   Experiments show that ASCC-defense outperforms the current state-of-the-arts in terms of robustness on two prevailing NLP tasks, \emph{i.e.}, sentiment analysis and natural language inference, concerning several attacks across multiple model architectures.
   Besides, we also envision a new class of defense towards robustness in NLP, 
   where our robustly trained word vectors can be plugged into a normally trained model and enforce its robustness without applying any other defense techniques.
   \footnote{Our code is available at https://github.com/dongxinshuai/ASCC.}
 \end{abstract}


 \section{Introduction}
  
 Recent extensive studies have shown that deep neural networks (DNNs) are vulnerable to adversarial attacks \citep{szegedy2013intriguing,goodfellow2014explaining,papernot2016limitations,kurakin2016adversarial,alzantot2018generating};
  \emph{e.g.}, minor phrase modification 
 can easily deceive Google’s toxic comment detection systems \citep{hosseini2017deceiving}. 
 This raises grand security challenges to advanced natural language processing (NLP) systems, 
 such as malware detection and spam filtering,
 where DNNs have been broadly deployed \citep{stringhini2010detecting,kolter2006learning}. 
 As a consequence, the research on defending against natural language
 adversarial attacks has attracted increasing attention.
 

 Existing  adversarial attacks in NLP can be categorized into three folds:
 \textit{(i)} character-level modifications \citep{belinkov2017synthetic,gao2018black,eger2019text},
 \textit{(ii)} deleting, adding, or swapping words \citep{liang2017deep,jia2017adversarial,iyyer2018adversarial},
 and \textit{(iii)} word substitutions using semantically similar words \citep{alzantot2018generating,ren2019generating,zang2020sememe}.
The first two attack types  usually break the grammaticality and naturality of the original input sentences, and thus can be detected by spell or grammar checker \citep{pruthi2019combating}.
In contrast, the third attack type only substitutes words with semantically similar words, 
thus preserves the syntactic and semantics of the original input to the most considerable extent
and are very hard to discern, even from a human's perspective. 
Therefore, building robustness against such word substitutions is a fundamental stepping stone towards robustness in NLP, which is the focus of this paper.

%
%

\begin{figure}[t]
   \centering
   \includegraphics[width=13.5cm]{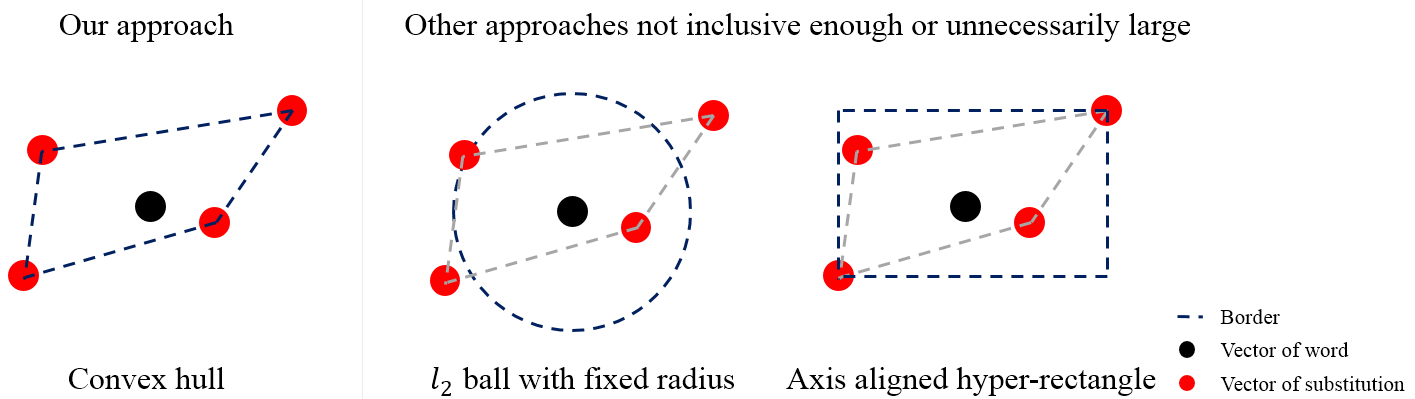}
   \vspace{-0em}
   \caption{Visualization of how different methods capture the word substitutions in the vector space.}\vspace{-0em}
   \label{fig:hull}
 \end{figure}

Adversarial attack by word substitution is a combinatorial optimization problem.
Solving this problem in the discrete textual space is considered NP-hard as the searching space increases exponentially with the length of the input. 
As such, many methods have been proposed to model word substitutions in the continuous word vector space \citep{sato2018interpretable, gong2018adversarial, jia2019certified, huang2019achieving}, 
so that they can leverage the gradients generated by a victim model either for attack or robust training.
However, previous methods capture word substitutions in the vector space by using either $l_2$-ball or hyper-rectangle, 
which results in perturbation sets that are not inclusive enough or unnecessarily large,
and thus impedes precise mimicry of the worst cases for robust training (see Fig.~\ref{fig:hull} for an illustration).
    
In this paper, we introduce a novel \textit{Adversarial Sparse Convex Combination} (ASCC) method, 
whose key idea is to model the solution space as a convex hull of word vectors.
Using a convex hull brings  two advantages: (\textit{i}) a continuous convex space is beneficial for gradient-based adversary generation, 
and (\textit{ii}) the convex hull is, by definition, the smallest convex set that contains all substitutions, thus is inclusive enough to cover all possible substitutions while ruling out unnecessary cases. 
In particular, we leverage a regularization term to encourage adversary towards an actual substitution, which aligns our modeling better with the discrete  textual space. 
We further propose ASCC-defense, which employs the ASCC to generate adversaries and incorporates adversarial training to gain robustness.
 
We evaluate ASCC-defense on two prevailing NLP tasks, \emph{i.e.}, sentiment analysis on IMDB and natural language inference on SNLI, across four model architectures, concerning two common attack methods.
Experimental results show that our method consistently yields models that are more robust than the state-of-the-arts with significant margins; \emph{e.g.}, we achieve $79.0\%$ accuracy under Genetic attacks on IMDB while the state-of-the-art performance is $75.0\%$.
Besides, our robustly trained word vectors can be easily plugged into standard NLP models and enforce robustness without applying any other defense techniques, which envisions a new class of approach towards NLP robustness.
For instance, using our pre-trained word vectors as initialization enhances a normal LSTM model to achieve $73.4\%$ robust accuracy,
while the state-of-the-art defense and the undefended model achieve $72.5\%$ and $7.9\%$, respectively.
 
\section{Preliminaries}
\label{sec:length}

 \subsection{Notations and Problem Setting}
 \label{subsec: setup}
 In this paper, we focus on text classification problem to introduce our method, while it can also be extended to other NLP tasks.
 We assume we are interested in training classifier $\mathcal{X}\rightarrow\mathcal{Y}$ that predicts label $y\in \mathcal{Y}$ given input $x \in \mathcal{X}$.
 The input $x$ is a textual sequence of $L$ words ${\{x_i\}}^{L}_{i=1}$.
 We consider the most common practice for NLP tasks where the first step is to map $x$
 into a sequence of vectors in a low-dimensional embedding space, which is denoted as $v(x)$.
 The classifier is then formulated as $p(y|v(x))$, 
 where $p$ can be parameterized by using a neural network, \emph{e.g.}, CNN or LSTM model.

We examine the robustness of a model against adversarial word substitutions \citep{alzantot2018generating,ren2019generating}. 
 Specifically, any word $x_{i}$ in $x$
 can be substituted with any word $\hat{x}_{i}$ in $\displaystyle \sS(x_{i})= {\{\displaystyle \sS(x_{i})_j\}}^{T}_{j=1}$,
 where $\displaystyle \sS(x_{i})$ represents a predefined substitution set for $x_{i}$ (including itself) and $T$ denotes the number of elements in $\displaystyle \sS(x_{i})$.
 To ensure that $\hat{x}$ is
likely to be grammatical and has the same label as $x$, $\displaystyle \sS(x_{i})$ is often comprised of semantically similar words of $x_{i}$, \emph{e.g.}, its synonyms.
 Attack algorithms such as Genetic attack \citep{alzantot2018generating} and PWWS attack \citep{ren2019generating} 
 aim to find the worst-case $\hat{x}$ to fool a victim model,
  whereas our defense methods aim to  build robustness against such substitutions.


 \subsection{Perturbation Set at Vector Level}
 \label{subsection: define perturbation}
 Gradients provide crucial information about a victim model for adversary generation \citep{szegedy2013intriguing,goodfellow2014explaining}.
 However, in NLP, the textual input space is neither continuous nor convex, which impedes effective use of gradients.
 Therefore, previous methods  
  capture perturbations in the vector space instead, by using the following simplexes (see Fig.\ref{fig:hull} for an illustration):

\textbf{$\boldsymbol{L_{2}}$-ball with a fixed radius.}
 \citet{miyato2016adversarial} first introduced adversarial training to NLP tasks.
 They use a $l_{2}$-ball with radius $\epsilon$ to constrain the perturbation, which is formulated as:
 \begin{align} 
   \hat{v}(x)=v(x)+r,~\mbox{s.t.}~ \| r  \|_2 \leq \epsilon, \label{eq:l2ad}
 \end{align}
 where $r$ denotes sequence-level perturbation in the word vector space and $\hat{v}$ denotes the adversarial sequence of word vectors.
 While such modeling initially considers $l_{2}$-ball at the sentence-level, it can also be extended to word-level to capture substitutions.
 Following that, \citet{sato2018interpretable} and \citet{barham2019interpretable} propose to additionally consider the directions
 towards each substitution.
 However, they still use the $l_{2}$-ball, 
 which often fails to capture the geometry of substitutions precisely.

 \textbf{Axis aligned bounds.}
 \citet{jia2019certified} and \citet{huang2019achieving} use  axis-aligned bound to capture  perturbations at the vector level.
 They consider the smallest axis-aligned hyper-rectangular that contains all possible substitutions.
 Such perturbation set provides useful properties for bound propagation towards robustness.
 However, the volume of the unnecessary space it captures can grow with the depth of the model and grow exponentially with the dimension of the word vector space.
 Thus it fits shallow architectures but often fails to  utilize the capacity of neural networks fully. 

 Besides, instead of fully defining the vector-level geometry of substitutions,
 \citet{ebrahimi2017hotflip} propose to find 
 substitutions by first-order approximation using directional gradients.
 It is  effective in bridging the gap between continuous embedding space and discrete textual space.
 However, 
 it is based on local approximation, which often fails to find global worst cases for robust training.

\section{Methodology}
 In this section, we first introduce the intuition of using a convex hull to capture substitutions.
 Then, we propose how Adversarial Sparse Convex Combination (ASCC) generates adversaries.
 Finally, we introduce ASCC-defense that incorporates adversarial training towards robustness.

 \subsection{Optimality of Using Convex Hull}
 
 From the perspective of adversarial defense,
  it is crucial to  well capture the attack space of word substitutions.
   There are three aspects we need to consider: 
 (i) \textit{Inclusiveness}: the space should include all  vectors of  allowed substitutions to cover all possible cases. 
 (ii) \textit{Exclusiveness}: on the basis of satisfying inclusiveness, 
 the space should be as small as possible since a loose set can generate unnecessarily intricate perturbations,
 which impede a model from learning useful information.
 (iii) \textit{Optimization}: the space should be convex and continuous to facilitate effective gradient-based optimization, whether the objective function is convex or not
 \citep{bertsekas1997nonlinear, jain2017non}.
 Inspired by archetypal analysis \citep{cutler1994archetypal}, 
 we propose to use a convex hull to build the attack space: the convex hull is a continuous space and, by definition, the minimal convex set containing all vectors of substitutions. 
 We argue that using a convex hull can satisfy all the above aspects (as illustrated in Fig.\ref{fig:hull}),
 and thus it  is   considered as theoretical optimum.

 
 \begin{figure*}[t]
  \centering
  \includegraphics[width=14cm]{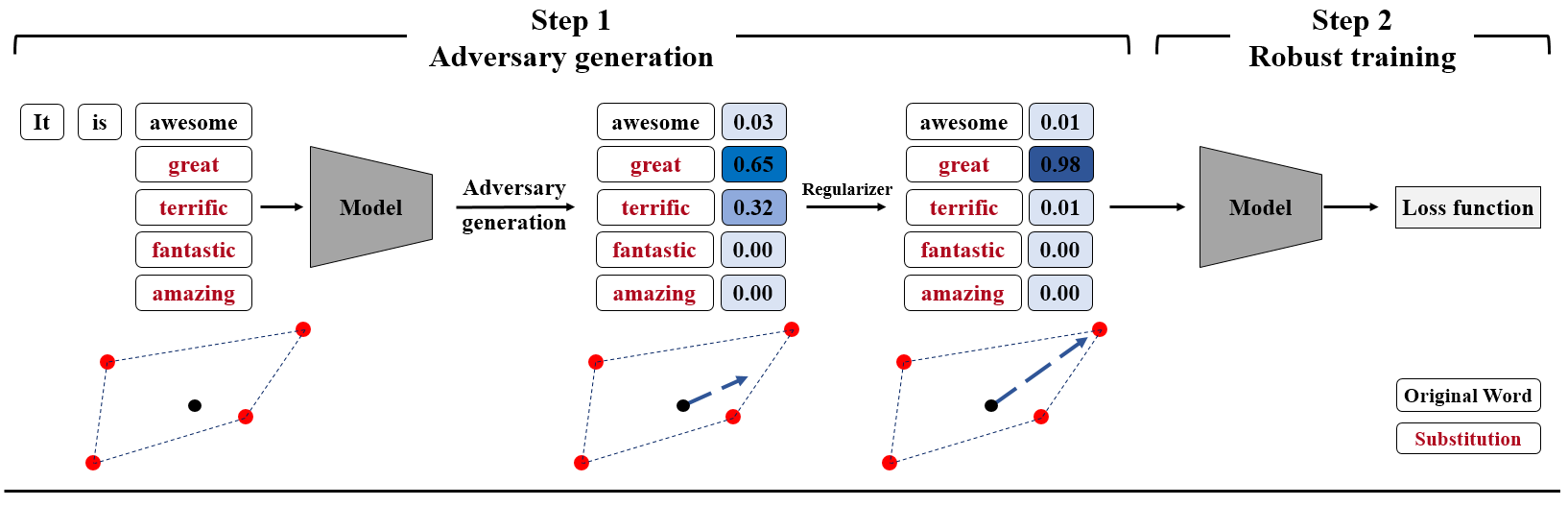}
  \vspace{-0em}
  \caption{An illustration of the training process of the ASCC-defense.
  Step 1: Generate adversaries by ASCC with regularization.
  Step 2: Take adversaries as input to perform adversarial training.
  }\vspace{-0em}
  \label{fig:procedure}
\end{figure*}

 \subsection{Adversarial Sparse Convex Combination}

 \textbf{Efficient representation of and optimization over a convex hull.}
 \label{subsection: generate adversarial}
 A classical workaround in literature for optimization over a constraint set is the projected gradient descent \citep{cauchy1847methode,frank1956algorithm,bubeck2014convex,madry2017towards}. 
  As for optimization over a convex hull, 
  it necessitates characterizing the convex hull, \emph{e.g.}, by vertexes, to perform projections.
  However, computing vertexes is computationally unfavorable because we need to recalculate the vertexes  
  whenever word embeddings change, which frequently occurs during the training process.
 
 In this paper, we propose a more efficient fashion for optimization over the concerning convex hull,
  based on the following proposition (the proof of which lies in the definition of convex hull):
 \begin{myproposition}\label{proposition:1}
   Let $\displaystyle \sS(u)=\{\displaystyle \sS(u)_{1},...,\displaystyle \sS(u)_{T}\}$ be the set of all substitutions of word $u$,
    $\textbf{\rm conv} \displaystyle \sS(u)$ be the convex hull of word vectors of all elements in $\displaystyle \sS(u)$, 
    and $v(\cdot)$ be the word vector function.
   Then, we have $\textbf{\rm conv} \displaystyle \sS(u) =  \{\sum\nolimits_{i=1}^{T} w_{i}v(\displaystyle \sS(u)_{i})~|~\sum\nolimits_{i=1}^{T} w_{i} = 1,w_{i} \geq 0 \}$.
 \end{myproposition}

 According to Proposition~\ref{proposition:1}, we can formulate $\hat{v}(x_{i})$, which denotes
 any vector in the convex hull around $v(x_{i})$, as:
 \begin{align}
  \hat{v}(x_{i})=\sum\nolimits_{j=1}^{T} w_{ij}v(\displaystyle \sS(x_{i})_{j}),~~\mathrm{s.t.}\sum\nolimits_{j=1}^{T} w_{ij} = 1,w_{ij} \geq 0. \label{eq:convex hull 1}
 \end{align}

As such, we use Eq.\ref{eq:convex hull 1} to transform the original optimization on $\hat{v}(x_{i})$ 
to the optimization on ${w_{i}}$, the coefficient of convex combination.
Considering that $w_{i}$ still belongs to a set with  constraint $\{\|w_{i}\|_{1}=1, w_{ij}\geq 0\}$,
to achieve better flexibilities of optimization, we introduce a variable $\hat{w}\in \mathbb{R}$ to 
relax the constraint on $w$ by the following equation:
 \begin{align}
   w_{ij}=\frac{\exp(\hat{w}_{ij})}{\sum_{j=1}^{T} \exp (\hat{w}_{ij})}, \hat{w}_{ij}\in\mathbb{R}. \label{eq:w}
 \end{align}
After such relaxation in Eqs.\ref{eq:convex hull 1}~and~\ref{eq:w}, 
we are able to optimize the objective function over the convex hull by optimizing   $\hat{w}\in \mathbb{R}$.
 It provides a projection-free way to generate any adversaries inside the convex hull using gradients. .

 \textbf{Gradient-based adversary generation.}
 Let $\mathcal{L}$ be a loss function concerning a classifier. 
 We can generate the worst-case convex combinations $\hat{v}(x)$ by finding the worst-case $\hat{w}$:
 \begin{align}
   \max\limits_{\hat{w}} \mathcal{L}(v(x), \hat{v}(x), y) \label{eq:acc}
 \end{align}where  $\mathcal{L}$ is classification-related, \emph{e.g.}, the cross-entropy loss over $\hat{v}(x)$: 
 \begin{align}
 \mathcal{L}(v(x), \hat{v}(x), y)=-\log p(y|\hat{v}(x)).
 \end{align}
 However, since we relax the discrete textual space to a convex hull in the vector space, 
 any $w_{i}$ that $\|w_{i}\|_{0} > 1$  is highly possible to give rise to $\hat{v}(x_{i})$ that does not correspond to a real substitution.
To align better  with  the discrete nature of textual input, we propose to impose a regularizer on the coefficient of convex combination, $w_{i}$. 
 To be specific, we take $w_{i}$ as a probability distribution and minimize the entropy function of $w_{i}$ to softly encourage the $l_{0}$ sparsity of $w_{i}$.
 We formulate this word-level entropy-based regularization term  as:
 \begin{align}
   \mathcal{H}(w_{i})= \sum\nolimits_{j=1}^{T} -w_{ij} \log(w_{ij}).
 \end{align}
Combining loss function $\mathcal{L}$ and the entropy-based regularizer $\mathcal{H}$,
we here formulate \textit{Adversarial Sparse Convex Combination} (ASCC) for adversary generation as:
 \begin{align}
   \max\limits_{\hat{w}} \mathcal{L}(v(x), \hat{v}(x), y) - \alpha \sum\nolimits_{i=1}^{L} \frac{1}{L} \mathcal{H}(w_{i}),\label{eq:ascc}
 \end{align}
 where $\alpha \geq 0$ is the weight controlling the regularization term (the effectiveness of which is validated in Sec.\ref{subsef:on the r}).


 \begin{algorithm}[t]
   \caption{ASCC-defense}
   \label{alg:algorithm}
   \textbf{Input}: dataset $\mathcal{D}$, parameters of Adam optimizer.\\
   \textbf{Output}: parameters $\theta$ and $\phi$.
   \begin{algorithmic}[1] 
   \REPEAT
     \FOR{random mini-batch  $\sim \mathcal{D}$}
       \FOR{ every $x$, $y$ in the mini-batch (in parallel)}
         \STATE  Solve the inner maximization in  Eq.\ref{eq:ascc train}  to find the optimal $\hat{w}$ by Adam;
         \STATE  Compute $\hat{v}(x)$ by Eq.\ref{eq:vh by wh} using $\hat{w}$  and then  compute the inner-maximum  in Eq.\ref{eq:ascc train};
       \ENDFOR
       \STATE   Update $\theta$ and $\phi$ by Adam to minimize the calculated inner-maximum;
     \ENDFOR
   \UNTIL{the training converges.}
   \end{algorithmic}
   \end{algorithm}\vspace{0em}
 
 \subsection{ASCC-Defense Towards Robustness}\vspace{-0em}
 \label{subsec:training}
 We here introduce ASCC-defense, which uses ASCC for adversaries and employs adversarial training towards robustness.
  We denote $\theta$ and $\phi$ as the parameters of $p(y|v(x))$ and $v(x)$, respectively.

 \textbf{Adversarial training paradigm for NLP.}
 Adversarial training \citep{szegedy2013intriguing,goodfellow2014explaining,madry2017towards}
 is currently one of the most effective ways to build robustness.
 \citet{miyato2016adversarial} are the first to use adversarial training for text classification.
 They use  $l_{2}$-ball with radius $\epsilon$ to restrict perturbations and
 the training objective can be defined as:
 \begin{align}
   \min\limits_{\theta,\phi} [\mathop{\mathbb{E}}\limits_{(x,y) \sim \mathcal{D}}[ \max\limits_{r} \mathcal{L}(v(x),\hat{v}(x), y,\theta,\phi) ]],~~~\mathrm{s.t.}~\hat{v}(x)=v(x)+r, \|r\|_{2}\leq \epsilon, \label{eq:ad train}
\end{align}
where $r$ denotes the perturbations in the vector space and $\mathcal{L}$ denotes a classification-related loss. 
Therefore, maximizing $\mathcal{L}$ can generate adversarial perturbations $r$ to fool a victim model, whereas minimizing $\mathcal{L}$ can let the model learn to predict under perturbations.

 \textbf{ASCC-Defense.}
 Instead of using $l_{2}$-ball in Eq.\ref{eq:ad train}, we leverage ASCC to capture perturbations inside the convex hull to perform adversarial training.
 This is to re-define $\hat{v}(x)$ in Eq.\ref{eq:ad train} using ASCC,
 and the resulting training objective is formulated as:
 \begin{align}
   &\min\limits_{\theta,\phi} [\mathop{\mathbb{E}}\limits_{(x,y) \sim \mathcal{D}}[ \max\limits_{\hat{w}} \mathcal{L}(v(x),\hat{v}(x), y,\theta,\phi) ]], \label{eq:ascc ad train}\\
   &\hat{v}(x_{i})=\sum\nolimits_{j=1}^{T} w_{ij}v(\displaystyle \sS(x_{i})_{j}),~w_{ij}=\frac{\exp(\hat{w}_{ij})}{\sum_{j=1}^{T} \exp (\hat{w}_{ij})}.\label{eq:vh by wh}
\end{align}

 To specify  $\mathcal{L}$ in Eq.\ref{eq:ascc ad train} for ASCC-defense,
  we consider the KL-divergence 
 between the prediction by vanilla input and the prediction under perturbations \citep{miyato2018virtual, Zhang2019theoretically}. 
  In the meantime, we also encourage the sparsity of $w_{i}$
 by the proposed regularizer for adversary generation. 
 Taking these together, we formulate the training objective of ASCC-defense as follows:
 \begin{align}\label{eq:ascc train}
   \min\limits_{\theta,\phi} [\mathop{\mathbb{E}}\limits_{(x,y) \sim \mathcal{D}}[ ~ \max\limits_{\hat{w}} -\log p(y|v(x)) - \alpha\sum_{i=1}^{L}  \frac{1}{L} \mathcal{H}(w_{i}) + \beta {\rm KL}(p(\cdot|v(x))||p(\cdot|\hat{v}(x))) ~]],
 \end{align}
 where $\alpha,\beta \geq 0$ control the weight of regularization and KL term, respectively.
 Noted that term $ \mathcal{H}(\cdot)$ has no gradient with respect to $\theta$ and $\phi$, so it only works during inner-max adversary generation.

 \textbf{Robust word vectors.}
 We here explain why ASCC-defense can yield more robust word vectors.
 Previous defenses such as \citet{miyato2016adversarial} fail to train word vectors in a robust way,
  as they update $\phi$ by only using the clean data flow.
Specifically,  \citet{miyato2016adversarial} obtains $\hat{v}(x_{i})$ through Eq.\ref{eq:l2ad},
 where perturbation $r$ has no gradient with respect to $\phi$, and thus $\nabla_{\phi} \hat{v}(x_{i}) = \nabla_{\phi} {v}(x_{i})$.
 On the contrary, $\hat{v}(x_{i})$ modeled by ASCC-defense has gradient \textit{w.r.t.} ${\phi}$
 concerning  all substitutions, as:
 \begin{align}
   \nabla_{\phi} \hat{v}(x_{i})= \sum\nolimits_{j=1}^{T} w_{ij}~\nabla_{\phi} v(S(x_{i})_{j}). \label{eq:gradients}
 \end{align}
 Therefore, ASCC-defense updates the word vector considering all potential adversarial substitutions simultaneously,
  which gives rise to  more robust word vectors
 (we validate our claim in Sec.\ref{subsec:exp robust word vector}).


 \begin{table}
     \caption{Accuracy(\%) of different defense methods under attacks on IMDB (\subref{tab:robustness on IMDB}) and SNLI (\subref{tab:robustness on SNLI}).
     ``First-order aprx'' denotes \citet{ebrahimi2017hotflip}.
     ``Adv $l_2$-ball'' denotes \citet{miyato2016adversarial}.
     ``Axis-aligned'' denotes \citet{jia2019certified}.
     ``ASCC-defense'' denotes the proposed method.
   }\vspace{-0em}
   \begin{subtable}{.5\linewidth}
   \begin{tabular}[t]{lccc}
   \hline
   \hline
   Method           & Model & Genetic                 & PWWS           \\
   \hline
   Standard         & LSTM         & 1.0                     & 0.2            \\ 
                    & CNN          & 7.0                     & 11.3           \\ 
   \hline
   First-order aprx & LSTM       & 72.5                     & 66.7          \\ 
                    & CNN          & 51.2                    & 74.1          \\ 
   \hline
   Adv $l_2$-ball   & LSTM         & 20.1                    & 11.7          \\ 
                    & CNN          & 36.7                    & 46.2          \\ 
   \hline
   Axis-aligned       & LSTM         & 64.7                    & 59.6        \\ 
                    & CNN          & 75.0                    &  69.5          \\ 
   \hline
   ASCC-defense     & LSTM         & \textbf{79.0}           & \textbf{77.1}  \\ 
                    & CNN          & \textbf{78.2}            &  \textbf{76.2}      \\ 
   \hline
   \hline
   \end{tabular}\vspace{-0em}
   \caption{Accuracy (\%) under attacks on IMDB.}\vspace{-0em}
   \label{tab:robustness on IMDB}
   \end{subtable}
   \begin{subtable}{.5\linewidth}
   \begin{tabular}[t]{lccc}
   \hline
   \hline
   Method           & Model        & Genetic             & PWWS          \\
   \hline
   Standard         & BOW          & 28.8                & 15.4          \\
                    & DCOM       & 30.2                & 9.0           \\
   \hline
   First-order aprx   & BOW        & 65.6                & 57.2         \\ 
                    & DCOM       & 66.7               & 58.6        \\ 
   \hline
   Adv $l_2$-ball        & BOW          & 35.0                & 16.7          \\ 
                    & DCOM       & 33.1                & 15.4          \\ 
   \hline
   Axis-aligned     & BOW          & 75.0                & 72.1             \\ 
                    & DCOM       & 73.7                & 67.9             \\ 
   \hline
   ASCC-defense     & BOW          & \textbf{76.3}       & \textbf{75.1} \\ 
                    & DCOM       &  \textbf{74.5}        & \textbf{72.8}  \\ 
   \hline
   \hline
   \end{tabular}\vspace{0em}
   \caption{Accuracy (\%) under attacks on SNLI.}\vspace{0em}
   \label{tab:robustness on SNLI}
   \end{subtable}

   \label{tab:main result}
   \end{table}

   \begin{figure}\vspace{0em}
     \begin{subfigure}{.50\textwidth}
       \centering
       \includegraphics[width=0.8\linewidth]{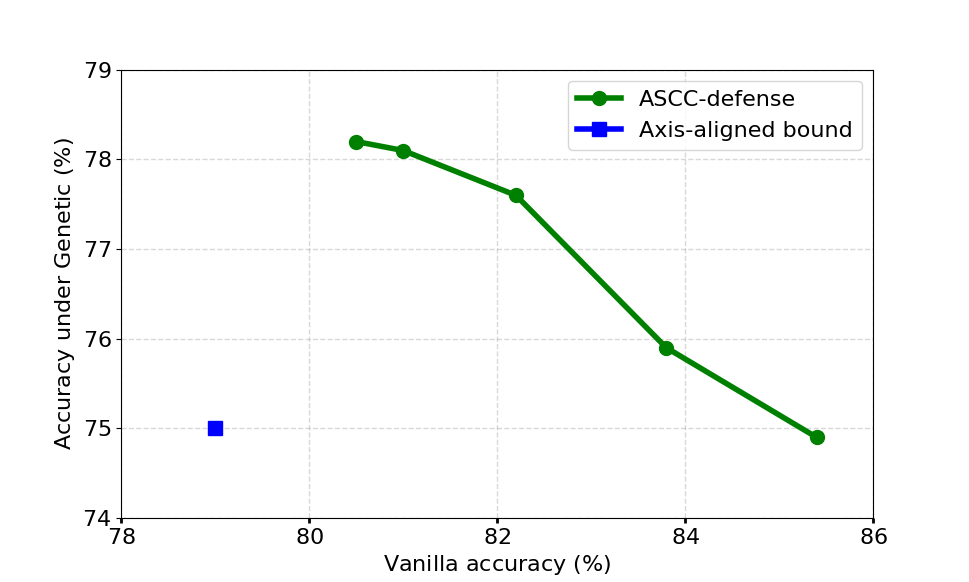}\vspace{0mm}
       \caption{Accuracy (\%) under Genetic attacks.}\vspace{-0em}
     \end{subfigure}
     \begin{subfigure}{.50\textwidth}
       \centering
       \includegraphics[width=0.8\linewidth]{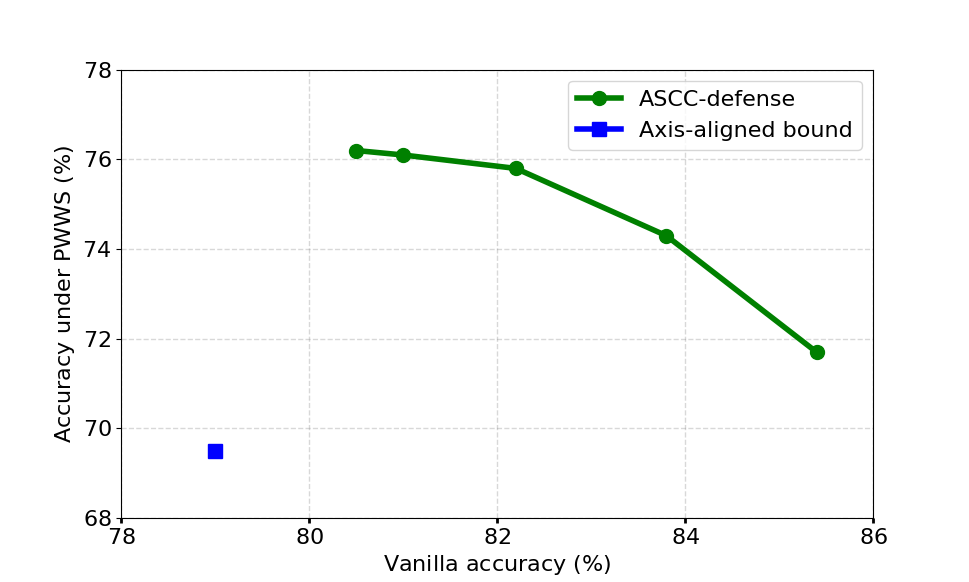}\vspace{-0mm}
       \caption{Accuracy (\%) under PWWS attacks.}\vspace{-0em}
     \end{subfigure}
     \caption{Tradeoff between robustness and accuracy on IMDB under Genetic and PWWS attacks. }\vspace{-0em}
     \label{fig:tradeoff}
   \end{figure}

 \textbf{Optimization.}
 We employ Adam \citep{kingma2014adam} to solve both inner-max and outer-min problems in Eq.\ref{eq:ascc train}.
 Our training process is illustrated in Fig.~\ref{fig:procedure} and presented  in Algorithm 1.

\section{Experiments}\vspace{-0em}
\label{sec:experiments}
 
 \subsection{Experimental Setting}\vspace{-0em}
 \noindent \textbf{Tasks and datasets.}
 We focus on two prevailing NLP tasks to evaluate the robustness and compare our method to the state-of-the-arts:
 (i) Sentiment analysis on the IMDB dataset \citep{maas-etal-2011-learning}.
 (ii) Natural language inference on the SNLI dataset \citep{bowman-etal-2015-large}.

 \noindent \textbf{Model architectures.}
 We examine robustness on the following four architectures to show the scalability of our method:
 (i) BOW, the bag-of-words model which sums up the word vectors and predicts by a multilayer perceptron (MLP),
 (ii) CNN model, 
 (iii) Bi-LSTM model,
 (iv) DCOM, decomposable attention model \citep{decomposable}, which generates context-aware vectors and predicts by a MLP.
 We align our implementation details with \citet{jia2019certified} for fair comparisons.

 \begin{table}[t]
   \caption{Ablation study on the sparsity regularization term.}\vspace{-0em}
   \begin{subtable}{.5\linewidth}
   \begin{tabular}[t]{lccc}
   \hline
   \hline
     Regul weight     &  Vanilla    & Genetic    & PWWS   \\
   \hline
       $\alpha$=0        &    80.6     &  75.1      &   61.6     \\
       $\alpha$=5        &   81.9      &  76.8      &   71.7 \\
       \textbf{$\boldsymbol{\alpha}$=10}       &    82.2     &  78.2      &   75.7   \\
       $\alpha$=15       &     81.2    &   76.1     &   78.3     \\
   \hline
   \hline
   \end{tabular}\vspace{-0em}
   \caption{Acc (\%) of CNN-based ASCC-defense on IMDB.}\vspace{-0em}
   \label{tab:robustness on IMDB}
   \end{subtable}
   \hspace{\fill}
   \begin{subtable}{.5\linewidth}
   \begin{tabular}[t]{lcccc}
   \hline
   \hline
    Regul weight    &  Vanilla    & Genetic    & PWWS   \\
   \hline
       $\alpha$=0        &       76.7  &     73.4   &   73.7 \\
       $\alpha$=5        &      77.4   &     75.1   &   74.0  \\
       \textbf{$\boldsymbol{\alpha}$=10}       &     77.8    &     76.3   &   75.1 \\
       $\alpha$=15       &    76.7     &     73.7   &   73.3 \\ 
   \hline
   \hline
   \end{tabular}\vspace{-0em}
   \caption{Acc (\%) of BOW-based ASCC-defense on SNLI.}\vspace{-0em}
   \label{tab:robustness on SNLI}
   \end{subtable}
   
   \bigskip 

   \label{tab:on alpha}
   \end{table}


 \noindent \textbf{Comparative methods.}
 (i) {Standard training.} It uses the cross-entropy loss as the main loss function. 
 (ii) {First-order approximation \citep{ebrahimi2017hotflip}.} While it is initially proposed to model char flip, it can also be applied to word substitutions.
  We implement its word-level version for comparison.
 (iii) {Adv ${l_2}$-ball \citep{miyato2016adversarial}.}  
 It first normalizes the word vectors and then generates  perturbations inside a $l_2$-ball for adversarial training.
 We implement it with word level $l_2$-ball and radius $\epsilon$ varying from $0.1$ to $1$.
 We only plot the best performance among different $\epsilon$.
 (iv) {Axis aligned bound \citep{jia2019certified}.} It models perturbations by an axis-aligned box and 
  uses bound propagation for robust training.
 (v) {ASCC-defense.} We set the hyper-parameters $\alpha=10$ and $\beta=4$.
 For fair comparisons,  KL term is employed for all compared adversarial training based methods. 
 More implementation details as well as runtime analysis can be found in the Appendix A.

 \begin{figure}\vspace{-0em}
  \begin{subfigure}{.50\textwidth}
    \centering
    \includegraphics[width=0.8\linewidth]{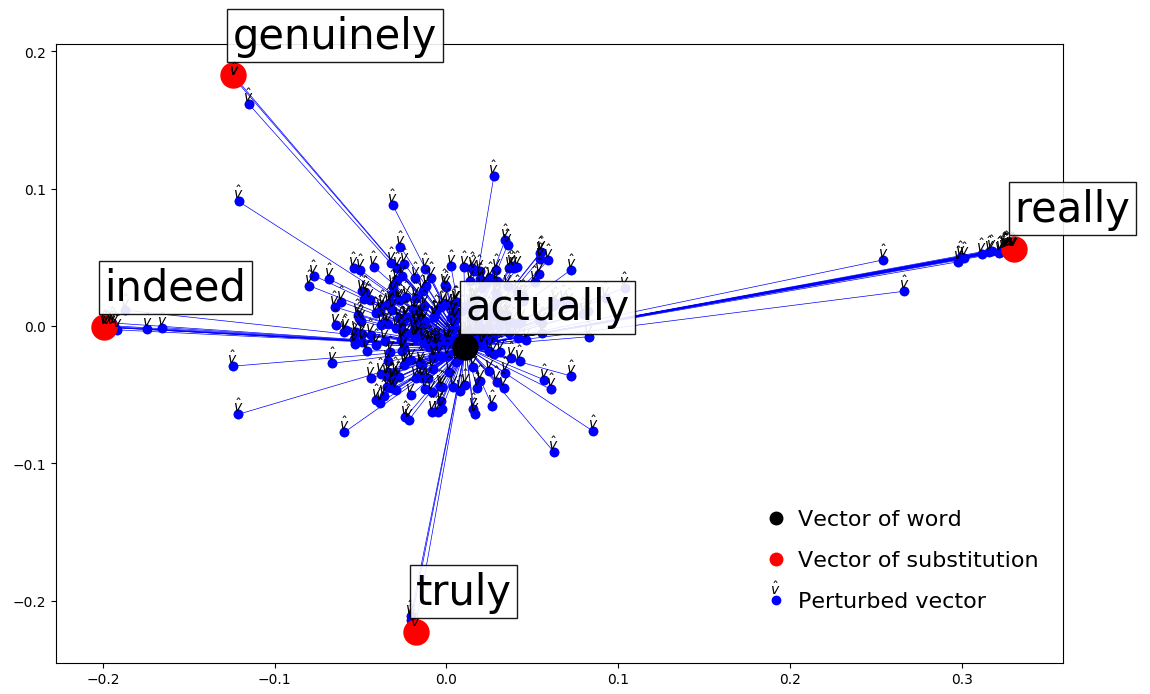}\vspace{-0em}
    \caption{$\hat{v}$ generated without the regularization term.}\vspace{-0em}
  \end{subfigure}
  \begin{subfigure}{.50\textwidth}
    \centering
    \includegraphics[width=0.8\linewidth]{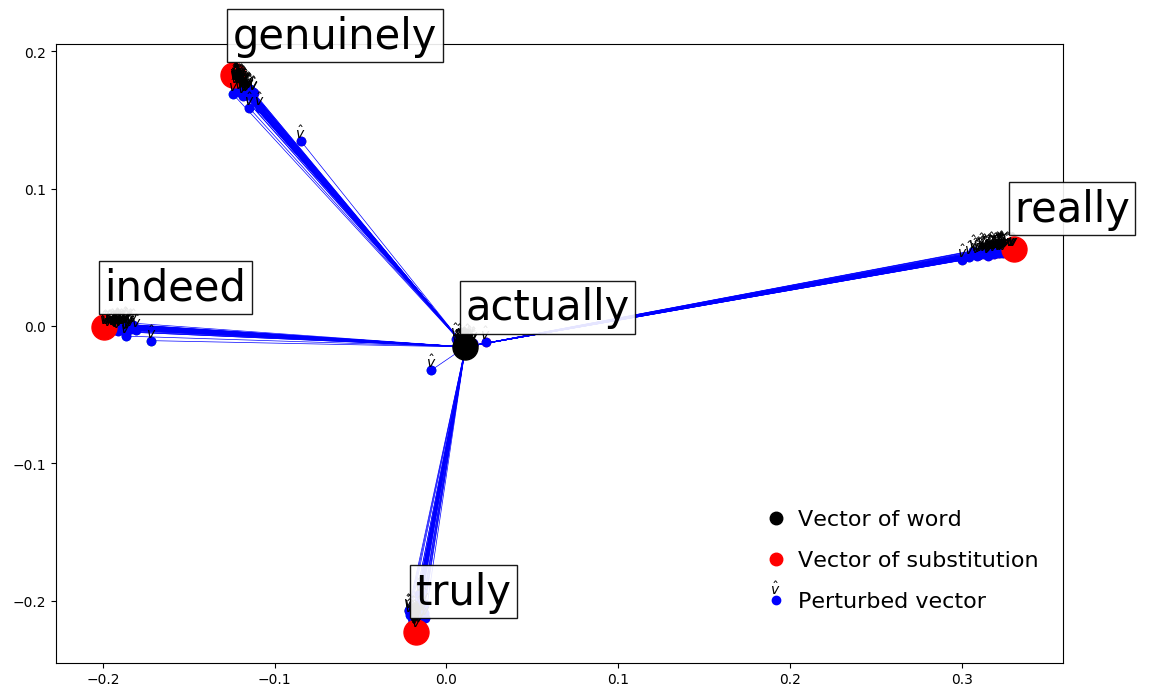}\vspace{-0em}
    \caption{$\hat{v}$ generated with the regularization when $\alpha=10$.}\vspace{-0em}
  \end{subfigure}
  \caption{An illustration to show the effectiveness of the proposed sparsity regularization.
  We randomly choose 300 adversaries of word ``actually'' from the test set of IMDB.
   Vectors are projected into a 2-dimensional space by SVD.
   Best view in color with zooming in.}\vspace{-0em} 
  \label{fig:on w}
\end{figure}

 \noindent \textbf{Attack algorithms.}
   We employ the following two powerful attack methods to examine the robustness:
   (i) Genetic attack \citep{alzantot2018generating} maintains a population to generate attacks in an evolving way.
   Aligned with \citet{jia2019certified}, we set the population size as $60$ to run for $40$ iterations.
   (ii) PWWS attack \citep{ren2019generating}  calculates the saliency of each word
    and then substitutes greedily.
    Aligned with \citet{alzantot2018generating} and \citet{jia2019certified}, we do not attack premise on SNLI.
 
 \noindent \textbf{Substitution set.}
 For fair comparisons with state-of-the-art  defense \citet{jia2019certified}, 
 we follow their setting to use substitution set from \citet{alzantot2018generating}.
 We apply the same language model constraint on Genetic as in \citet{jia2019certified},
  while do not apply it on PWWS attacks.
  As we aim at using prevailing setting to compare with the state-of-the-arts, 
   we do not focus on how to construct the substitution set in this work and leave it for future exploration.

 \subsection{Main Result}
 Aligned with \citet{jia2019certified}, we evaluate robustness on 1000 randomly selected examples from the test set of IMDB,
  and all 9824 test examples from SNLI. 
 As shown in Tab.\ref{tab:main result}, our method achieves leading robustness across all architectures with significant margins.
 For example, on IMDB, we surpass LSTM-based runner-up method by $6.5\%$ under Genetic and $10.4\%$ under PWWS attacks.
 
 Plus, the robust performance of ASCC-defense is consistent against different attacks: 
 \emph{e.g.}, on IMDB, LSTM-based ASCC-defense achieves $79.0\%$ under Genetic attacks and $77.1\%$ under PWWS attacks,
  which shows ASCC-defense does not rely on over-fitting to a specific attack algorithm.
 In addition to robust accuracy, we also plot the tradeoff between robustness and accuracy  in Fig.\ref{fig:tradeoff}, 
 which shows our method can trade off some accuracy for more robustness compared to the state-of-the-art.
 More detailed vanilla accuracy and our performance on BERT \citep{devlin2018bert} are shown in Appendix B.

 \begin{table}[t]
   \caption{Accuracy (\%) of models  initialized with different word vectors \textit{without} any other defense technique.
   GloVe denotes the word vectors from \citet{pennington2014glove}. 
   ``First order V'' denotes word vectors trained by \citet{ebrahimi2017hotflip}.
   ``ASCC-V'' denotes word vectors trained by ASCC-defense. We freeze the pre-trained word vectors during normal training.
   }\vspace{-0em}
   \begin{subtable}{.5\linewidth}
   \begin{tabular}[t]{lcccc}
   \hline
   \hline
   Word vector        & Model       & Vanilla & Genetic                   \\ 
   \hline
   GloVe           & LSTM      & \textbf{88.5} & 7.9                            \\
   First order V   &LSTM       & 85.3  & 65.6                         \\
   ASCC-V          &LSTM       & 84.1 & \textbf{73.4}             \\
   \hline
   GloVe            & CNN       & \textbf{86.4}  & 8.6                             \\
   First order V    & CNN       &  83.1  & 44.7                          \\
   ASCC-V           & CNN       & 84.2   & \textbf{72.0}            \\
   \hline
   \hline
   \end{tabular}\vspace{-0em}
   \caption{Accuracy (\%) under attacks on IMDB.}\vspace{-0em}
   \label{tab:robustness on IMDB}
   \end{subtable}
   \hspace{\fill}
   \begin{subtable}{.5\linewidth}
   \begin{tabular}[t]{lccc}
   \hline
   \hline
   Word vector        & Model  & Vanilla & Genetic              \\ 
   \hline
   GloVe             & BOW    & \textbf{80.1}& 35.8                       \\
   First order V     & BOW    & 79.3& 62.1                      \\
   ASCC-V           & BOW     & 77.9& \textbf{69.6}          \\
   \hline
   GloVe            & DCOMP   & \textbf{82.6}& 41.8                          \\
   First order V    & DCOMP   & 78.7 & 62.8                         \\
   ASCC-V           & DCOMP   & 77.8 & \textbf{72.1}           \\
   \hline
   \hline
   \end{tabular}\vspace{-0em}
   \caption{Accuracy (\%) under attacks on SNLI.}\vspace{-0em}
   \label{tab:robustness on SNLI}
   \end{subtable}
   \label{tab:on embedding}
   \end{table}

   \subsection{On the Regularization and Other Discussions}
   \label{subsef:on the r}
   Fig.\ref{fig:on w} qualitatively shows how the proposed regularizer encourages sparsity.
   After applying the regularization, 
   the resulting $\hat{v}$ is close to a substitution,
   corresponding better with the discrete nature of  textual input.
   Tab.\ref{tab:on alpha} quantitatively shows the influence of the regularization term on robustness.
   Specifically, when $\alpha=10$ our method performs the best.
   As $\alpha$ keeps increasing, ASSC focus too much on the sparsity and thus fail to find strong enough perturbations for  robust training.

   We now discuss some other intuitive defense methods.
   The thought of enumerating all combinations during training is natural and yet impractical on benchmark datasets;
   \emph{e.g.}, on IMDB 
   the average number of combinations per input is $6^{108}$. 
   Augmenting training with random combinations is also ineffective, 
   since it fails to find hard cases in the exponentially large attack space;
   \emph{e.g.}, under Genetic ASCC-defense surpasses random augmentation by 46.0\% on IMDB and by 7.8\% on SNLI 
   (more significant margin owes to larger attack space on IMDB).
   Besides, though simply grouping all substitutions can achieve ensured robustness,
    it sacrifices discriminative powerness:
    two words that are not semantically similar will be mapped together
    just because they are indirectly related by one or more mediators.
   For instance,
   grouping defense achieves 71.3\% robust accuracy and 71.3\% vanilla accuracy on IMDB
   while ASCC-defense achieves 79.0\% and 82.5\% respectively.



 \subsection{Robust Word Vectors}
 \label{subsec:exp robust word vector}
 As mentioned in Sec.\ref{subsec:training}, ASCC-defense updates the vector of a word  by considering all its substitutions,
 and thus the obtained word vectors are robust in nature.
 To validate, we use the standard training process to train models but with different pre-trained word vectors as initialization.
 We compare  word vectors pre-trained by ASCC-defense  with Glove and word vectors pre-trained by \citet{ebrahimi2017hotflip}
 (the best performing setting for \citet{miyato2016adversarial} and \citet{jia2019certified} is to freeze the word vectors as Glove, 
 which laterally validates our claim about robust word vectors in Sec.\ref{subsec:training}). 
 As shown in Tab.\ref{tab:on embedding}, the models initialized by our robustly trained word vectors (and fixed during normal training)
 are robust to attacks without applying any other defense techniques.
 For example, armed with our robust word vectors, a normally trained LSTM-based model can achieve $73.4\%$ under Genetic attacks,
 whereas using GloVe  achieves $7.9\%$.

 In addition, this result also implies a new perspective towards robustness in NLP:
 the vulnerabilities of NLP models relate to  word vectors significantly, 
 and transferring pre-trained robust word vectors can be a more scalable way towards NLP robustness.
  For more result, please refer to Appendix B. 

\section{Related Work}\vspace{0em}

Though achieved success in many fields, DNNs appear to be susceptible to adversarial examples \citep{szegedy2013intriguing}. 
Initially introduced to attack CV models, attack algorithms vary from ${L_p}$ bounded \cite{goodfellow2014explaining,carlini2017towards,madry2017towards},
 universal perturbations \citep{moosavi2017universal,liu2019universal},
to wasserstein distance-based attack \citep{wong2019wasserstein}, while defense techniques for CV models 
include adversarial training\citep{goodfellow2014explaining,kurakin2016adversarial,madry2017towards}, preprocessing \citep{chen2018improving,yang2019menet}, 
and generative classifiers \citep{li2018generative,schott2018towards,dong2020api}.

Recently, various classes of NLP adversarial attacks have been proposed.
Typical methods consider char-level manipulations 
\citep{hosseini2017deceiving,ebrahimi2017hotflip,belinkov2017synthetic, gao2018black,eger2019text,pruthi2019combating}.
Another line of thought focus on deleting, adding, or swapping words 
 \citep{iyyer2018adversarial,ribeiro2018semantically,jia2017adversarial,zhao2017generating}.

In contrast to char-level and sequence-level manipulations, word substitutions consider prior knowledge  to preserve semantics and syntactics,
such as synonyms from WordNet \citep{miller1998wordnet}, 
Sememes \citep{bloomfield1926set,dong2006hownet}, and neighborhood relationships \citep{alzantot2018generating},
Some focus on heuristic searching in the textual space \citep{alzantot2018generating,liang2017deep,ren2019generating,jin2019bert,zhang2019generating,zang2020sememe},
while \citep{papernot2016crafting,gong2018adversarial} propose to leverage gradients for adversary generation  in the vector space.

As for defense against adversarial word substitutions, 
\citet{ebrahimi2017hotflip} find substitutions  by first-order approximation.
\citet{miyato2016adversarial}, \citet{barham2019interpretable} and \citet{sato2018interpretable} use $l_2$-ball to model perturbations,
while 
\citet{jia2019certified} and \citet{huang2019achieving} use axis-aligned bound.
\citet{zhou2020defense}  sample from Dirichlet distribution to initialize a convex combination of substitutions, but the sparsity might be lost during adversary generation.
Our work differs
as we model the convex hull with sparsity by entropy function, and the sparsity is enforced during the whole process,
which makes our  captured  geometry of substitutions more precise.
%
 %


\section{Conclusion}\vspace{-0em}

In this paper, we proposed a novel method to use the convex hull to capture and defense against adversarial word substitutions.
Our method yields models that consistently surpass the state-of-the-arts across datasets and architectures.
The experimental results further demonstrated that the word vectors themselves can be  vulnerable and 
our method gives rise to robust word vectors that can enforce robustness  without applying any other defense techniques.
As such, we hope this work can be a stepping stone towards even broader  robustness in NLP.

\section*{Acknowledgements}

This work is supported by the National Science Fund for Distinguished Young Scholars (No.~62025603),
 and the National Natural Science Foundation of China (No.~U1705262, No.~62072386, No.~62072387, No.~62072389, No.~62002305, No.~61772443, No.~61802324, and No.~61702136).


\bibliography{iclr2021_conference}
\bibliographystyle{iclr2021_conference}

\appendix

\setcounter{table}{3}
\setcounter{figure}{4}
\setcounter{equation}{13}

\section{Appendix}

\subsection{Implementation Details}\vspace{-0em}
\textbf{Text processing.}
We employ the tokenizer from NLTK and ignore all punctuation marks when tokenizing the textual input.
We set the maximum length of input as $300$ for IMDB and $80$ for SNLI. For unknown words, we set them as "NULL" token.

\textbf{Hyper-parameters and optimization.}
We set $\alpha$ as $10$ and $\beta$ as $4$ for the training procedure defined in Eq.12. 
To generate adversaries for robust training,
we employ Adam optimizer with a learning rate of $10$ and a weight decay of $0.00002$ to run for $10$ iterations 
To update $\phi$ and $\theta$, we also employ Adam optimizer, 
the parameters of which differ between architectures and will be discussed as follows.

\textbf{Architecture parameters}
(i) CNN for IMDB: 
We use a 1-d convolutional layer with kernal size of 3 to extract features and then make predictions.
We set the batch-size as 64 and use Adam optimizer with a learning rate of $0.005$ and a weight decay of $0.0002$.
(ii) Bi-LSTM for IMDB: 
We use a bi-directional LSTM layer to process the input sequence, and then use the last hidden state to make predictions.
We set the batch-size as 64 and use Adam optimizer with a learning rate of $0.005$ and a weight decay of $0.0002$.
(iii) BOW for SNLI: We first sum up the word vectors at the dimension of sequence  and concat the encoding of the premise and the hypothesis.
Then we employ a MLP of 3 layers to predict the label.
We set the batch-size as 512 and use Adam optimizer with a learning rate of $0.0005$ and a weight decay of $0.0002$.
(iv) DECOMPATTN for SNLI: We first generates context-aware vectors 
and then employ a MLP of 2 layers to make predictions given the context-aware vectors. 
We set the batch-size as 256 and use Adam with a learning rate of $0.0005$ and a weight decay of $0$.

\subsection{Runtime Analysis}

All models are trained using the GeForce GTX1080 GPU.
(i)
As for IMDB, it takes about $1.5$ GPU hours to train a CNN-based model and $2$ GPU hours for  a LSTM-based model.
(ii)
As for SNLI, it takes about $12$ GPU hours to train a BOW-based model and $15$ GPU hours for DECOMPATTN-based model.


  \section{Additional Experimental Result}

  \subsection{Vanilla Accuracy and Robust Accuracy under Genetic attacks without Constraint}\vspace{-0.5em}
  In Tab.\ref{tab:main result}, we have plotted the robust accuracy under attacks to compare with state-of-the-arts.
  Here we plot the vanilla accuracy of all compared methods in Tab.\ref{tab:vanilla accuracy} as an addition.
  Aligned  with Tab.\ref{fig:tradeoff}, the parameters of each compared method here are chosen to have the best robust accuracy instead of vanilla accuracy.

  In our main result in Tab.\ref{tab:main result}, for fair comparisons we align our setting with SOTA defense Jia et al.,
   where genetic attacks are constrained by a language model.
  Here we  plot our performance under genetic attacks without any language model constraint as an addition:
  ASCC-defense achieves 76.7\% robust accuracy under Genetic without constraint on IMDB based on LSTM, and 72.8\% on SNLI based on BOW.

  \begin{table}[t]
      \caption{Vanilla accuracy(\%) of different defense methods on IMDB (\subref{tab:robustness on IMDB}) and SNLI (\subref{tab:robustness on SNLI}).
    }\vspace{-0.5em}
    \begin{subtable}{.5\linewidth}
    \begin{tabular}[t]{lcc}
    \hline
    \hline
    Method           & Model & Vanilla accuracy             \\    
    \hline
    Standard         & LSTM         & 88.5                      \\ 
                     & CNN          & 87.2                    \\ 
    \hline
    First-order aprx & LSTM         & 83.2                     \\ 
                     & CNN          & 80.3                   \\ 
    \hline
    Adv $l_2$-ball   & LSTM         & 84.6                   \\ 
                     & CNN          & 84.5                   \\ 
    \hline
    Axis-aligned     & LSTM         & 76.8                   \\ 
                     & CNN          & 81.0                     \\ 
    \hline
    ASCC-defense     & LSTM         & 82.5          \\ 
                     & CNN          & 81.7          \\ 
    \hline
    \hline
    \end{tabular}\vspace{-0em}
    \caption{Vanilla accuracy (\%) on IMDB.}\vspace{-0em}
    \label{tab:robustness on IMDB}
    \end{subtable}
    \begin{subtable}{.5\linewidth}
    \begin{tabular}[t]{lcc}
    \hline
    \hline
    Method           & Model        & Vanilla accuracy         \\
    \hline
    Standard         & BOW          & 80.1                \\
                     & DECOMP       & 82.6               \\
    \hline
    First-order aprx & BOW          & 78.2              \\ 
                     & DECOMP       & 77.6                \\ 
    \hline
    Adv $l_2$-ball   & BOW          & 74.8             \\ 
                     & DECOMP       & 73.5              \\ 
    \hline
    Axis-aligned     & BOW          & 79.4              \\ 
                     & DECOMP       & 77.1              \\ 
    \hline
    ASCC-defense     & BOW          & 77.2       \\ 
                     & DECOMP       & 76.3       \\ 
    \hline
    \hline
    \end{tabular}\vspace{0em}
    \caption{Vanilla accuracy (\%)  on SNLI.}\vspace{0em}
    \label{tab:robustness on SNLI}
    \end{subtable}
 
    \label{tab:vanilla accuracy}
    \end{table}

    \begin{table}[t]\vspace{-1em}
      \centering
        \caption{Vanilla and robust accuracy (\%) of the proposed method on BERT (bert-base-uncased).}
      \begin{tabular}[t]{lcccc}
        \hline
        \hline
      Method           & Dataset & Model & Vanilla accuracy    &  Under Genetic attack         \\    
      \hline
      Standard         & IMDB & BERT    & 92.2       &     16.4         \\ 
       ASCC-defense    & IMDB & BERT    & 77.5        &      70.2      \\ 
      \hline
      \hline
      \end{tabular}
     
      \label{tab:bert}
      \end{table}

\begin{table}[]
  \caption{Accuracy (\%) of normally trained models initialized with and freezed by the proposed robust word vectors.
  For example, pre-trained on LSTM means the word vectors are pre-trained by LSTM-based ASCC-defense and 
  applied to CNN means the pre-trained word vectors are used to initialize a CNN model to perform normal training.
  }\vspace{-0.5em}
  \begin{subtable}{.5\linewidth}
  \begin{tabular}[t]{lccc}
  \hline
  \hline
  Pre-trained & Applied to          & Vanilla        & Genetic                          \\ 
  \hline
  LSTM&LSTM  & 84.1           & {73.4}                  \\
  LSTM&CNN  & 78.5           & 71.9                           \\
  \hline
  \hline
  \end{tabular}\vspace{-0em}
  \caption{Accuracy (\%) under attacks on IMDB.}\vspace{0em}
  \label{tab:robustness on IMDB}
  \end{subtable}
  \hspace{\fill}
  \begin{subtable}{.5\linewidth}
  \begin{tabular}[t]{lccc}
  \hline
  \hline
  Pre-trained & Applied to        & Vanilla        & Genetic                       \\ 
  \hline
  DECOMP&DECOMP  & 77.8           & {72.1}                   \\
  DECOMP&BOW     & 77.2           & 70.5                        \\
  \hline
  \hline
  \end{tabular}\vspace{-0em}
  \caption{Accuracy (\%) under attacks on SNLI.}\vspace{0em}
  \label{tab:robustness on SNLI}
  \end{subtable}
  
  \bigskip 

  \label{tab:cross archi rv}
  \end{table}\vspace{-0em}

  \subsection{Performance on BERT}
  ASCC-defense models perturbations at word-vecotr level, and thus it can be applied to architectures like Transformers, as long as it uses word embeding as its first layer.
  To validate, we conduct experiments on BERT (bert-base-uncased) using standard training and ASCC-defense respectively.
  As shown in Tab.\ref{tab:bert}, ASCC-defense   enhances the robustness of BERT model significantly.
  Specifically, BERT finetuned by standard method on IMDB achieves 16.4\% robust accuracy under Genetic attacks, 
  while using the proposed ASCC-defense achieves 70.2\%.

  \subsection{Reduced Perturbation Region}
  In this section, we show the reduced perturbation region by  using convex hull compared to $l_2$-ball and hyper-rectangle.
  To make the result more intuitive, we first project word vectors into a 2D space by SVD, 
  and than calculate the average area of each modeling (to rule out irrelevant factors, we use word vectors from  GloVe and consider words whose substitution sets are of the same size).
  We choose the smallest $l_2$-ball and hyper-rectangle that contain all substitutions to compare with convex hull.
  The result shows that using convex hull reduce the perturbation region significantly.
  Specificaly, the average ratio of the area modeled by convex hull to the area modeled by hyper-rectangle  is 29.2\%,
   and the average ratio of the area modeled by convex hull  to the area modeled by $l_2$-ball area is 8.4\%.

  \subsection{Cross-Architectures Performance of Robust Word Vectors}
  As discussed in Sec.4.4, 
  our robustly trained word vectors can enforce the robustness of a normally trained model without applying any other defense techniques.
  In this section, we aim to examine whether such a gain of robustness over-fits to a specific architecture.
  To this end, we first employ ASCC-defense to obtain robust word vectors and 
  then fix the word vectors as the initialization of 
  another model based on a different architecture to perform normal training.
  We examine the accuracy  under attacks.
  As shown in Tab.\ref{tab:cross archi rv}, our robustly trained word vectors consistently enhance the robustness of a normally trained model
   based on different architectures.
  For example, though trained by a LSTM-based model, the robust word vectors can still enforce 
  a CNN-based model to achieve robust accuracy of $71.9\%$ under Genetic attacks (whereas initializing by GloVe achieves $8.6\%$),
  demonstrating the across-architecture transferability of the robustness of our pre-trained word vectors .

\end{document}

%% file: math_commands.tex
\usepackage{amsmath,amsfonts,bm}

\def\eqref#1{equation~\ref{#1}}

\def\1{\bm{1}}

\DeclareMathAlphabet{\mathsfit}{\encodingdefault}{\sfdefault}{m}{sl}
\SetMathAlphabet{\mathsfit}{bold}{\encodingdefault}{\sfdefault}{bx}{n}

\def\sS{{\mathbb{S}}}

